# The Basic Kak Neural Network with Complex Inputs

Pritam Rajagopal

The Kak family of neural networks [3-6,12] is able to learn patterns quickly, and this speed of learning can be a decisive advantage over other competing models, such as backpropagation networks, perceptrons or generalized Hopfield networks [7,8], in many applications. Amongst the many implementations of these networks are those using reconfigurable networks, FPGAs and optical networks [2,11,13].

In some applications, it is useful to use complex data [1], and it is with that in mind that I present an introduction to the basic Kak network with complex inputs. This network for complex inputs uses the 3C algorithm that was published before in a book chapter [10]; the purpose of this article is to add further observations and place the work in appropriate contextual framework.

This algorithm is of prescriptive learning, where the network weights are assigned simply upon examining the inputs. The performance of the algorithm was tested using the pattern classification experiment and the time series prediction experiment with the Mackey-Glass time series. An input encoding called quaternary encoding is used for both experiments since it reduces the network size significantly by cutting down on the number of neurons that are required at the input layer.

The Kak network family is part of a larger hierarchy of learning schemes that include quantum models [14]. The quantum models themselves come with much promise and matching problems [15-20]. The larger issue of network models that can equal biological learning will not be taken up here. But, in a sense, networks with complex inputs are part of the hierarchy of models and, therefore, they have scientific implications that go beyond their narrow engineering use.

## 1 Introduction

The training processes used for different architectures are iterative, requiring a large amount of computer resources and time for the training. This may not be desirable in some applications. The Kak family of networks (using algorithms CC1 to CC4) [3-9] speeds up the training process of neural networks that handle binary inputs, achieving instantaneous training. A version for mapping non-binary inputs to non-binary outputs also exists [12].

The corner classification approach utilizes prescriptive learning. In this procedure, the network interconnection weights are assigned based entirely on the inputs without any computation. The corner classification algorithms such as

CC3 and CC4 are based on two main ideas that enable the learning and generalization of inputs:

1. The training vectors are mapped to the corners of a multidimensional cube. Each corner is isolated and associated with a neuron in the hidden layer of the network. The outputs of these hidden neurons are combined to produce the target output.

2. Generalization using the *radius of generalization* enables the classification of any input vector within a Hamming Distance from a stored vector as belonging to the same class as the stored vector.

Due to its generalization property, the CC4 algorithm can be used efficiently for certain AI problems. When sample points from a pattern are presented to the network, the CC4 algorithm trains it to store these samples. The network then classifies the other input points based on the radius of generalization, allowing for the network to recognize the pattern with good accuracy. In time-series prediction, some samples from the series are used for training, and then the network can predict future values in the series.

Here the generalization of the corner classification is used to handle complex inputs, which uses a new procedure of weight assignment. The next section describes an encoding scheme called the quaternary encoding, which will be used in different experiments to analyze the new algorithm. Section 3 presents the 3C algorithm, and in section 4 the performance of the algorithm is tested using the time series prediction experiment. Finally, the last section provides the conclusions related to the use of complex binary inputs in corner classification and the future of the 3C algorithm.

## 2    Quaternary Encoding

The quaternary encoding scheme is a simple modification of the unary scheme and accommodates two additional characters *i* and *1+i* besides 0 and 1. Due to the additional characters in this scheme, the length of the *codewords* for a range of integers is reduced when compared to unary. For example the integers 1 to 16 is represented using quaternary codewords only five characters, whereas the unary required 16-bit strings for the same range of numbers. Table 1 shows the set of codewords used to represent the integers 1 to 16.

**Length of the codewords**

An important issue is to decide the length of the codewords required to represent a desired range of integers. Let $l$ be the length of the codewords for a range of $C$ integers. Consider the integers in Table 1. For this range $C = 16$ and $l = 5$. We can now examine how 16 codewords can be formed with $l = 5$. The 16 codewords can be classified into three groups. The first group represents integers 1 to 6, where the codewords are constructed without using characters *i* or *1+i*.

The codewords in the second group represent integers 7 to 11 and don't use *1+i*, while in the third group the codewords representing integers 12 to 16 use *1+i*.

Table 1: Quaternary codewords for integers 1 to 16

| Integer | Quaternary code | | | | |
|---|---|---|---|---|---|
| 1 | 0 | 0 | 0 | 0 | 0 |
| 2 | 0 | 0 | 0 | 0 | 1 |
| 3 | 0 | 0 | 0 | 1 | 1 |
| 4 | 0 | 0 | 1 | 1 | 1 |
| 5 | 0 | 1 | 1 | 1 | 1 |
| 6 | 1 | 1 | 1 | 1 | 1 |
| 7 | 1 | 1 | 1 | 1 | i |
| 8 | 1 | 1 | 1 | i | i |
| 9 | 1 | 1 | i | i | i |
| 10 | 1 | i | i | i | i |
| 11 | i | i | i | i | i |
| 12 | i | i | i | i | 1+i |
| 13 | i | i | i | 1+i | 1+i |
| 14 | i | i | 1+i | 1+i | 1+i |
| 15 | i | 1+i | 1+i | 1+i | 1+i |
| 16 | 1+i | 1+i | 1+i | 1+i | 1+i |

We see here that the first group has 6 codewords. The other two have 5 each, corresponding to the length of the codewords as the next new character fills up one position after another in each successive codeword. For any *C*, the set of codewords would consist of three such groups where the first group has $l + 1$ codewords, and the second and third have *l* codewords each. This can be summarized as follows:

$$C = (l + 1) + l + l \qquad (1)$$
$$C = 3 * l + 1 \qquad (2)$$
$$l = (C - 1) / 3 \qquad (3)$$

Equation 3 is valid only when (*C* – 1) is divisible by 3. For cases when this is not true, we obtain:

$$l = \textbf{ceil } [(C - 1) / 3] \qquad (4)$$

When (*C* – 1) is not divisible by 3, the number of codewords that can be formed using the *l* obtained from Equation 4 is more than required. In this case any *C* consecutive codewords from the complete set of words of length *l* may be used.

# 3   The 3C algorithm

The 3C algorithm (from Complex Corner Classification, CCC) is a generalization of the CC4 and is capable of training 3-layered feedforward networks to map inputs from the alphabet {0, 1, i, 1+i} to the real binary outputs 0 and 1. This algorithm uses a different procedure for the assignment of the input interconnection weights when compared to the CC4 algorithm. Therefore the combination procedure of these weights with the inputs is also different. The features of the algorithm and its network are:

1. The number of input neurons is one more than the number of input elements in a training sample. The extra neuron is the bias neuron which is always set to one.

2. A hidden neuron is created for each training sample; the first hidden neuron corresponds to the first training sample, the second neuron corresponds to the second sample and so on.

3. The output layer is fully connected; each hidden neuron is connected to all the output neurons.

4. The interconnection weights from all the input neurons excluding the bias neuron are complex. Each input of the alphabet {0, 1, i, 1+i} is treated as complex for the weight assignment.

5. If the real part of the input element is 0 then the real part of the corresponding input interconnection weight is assigned as -1. If the real part is of the input element is 1 then the real part of the weight is also set as 1.

6. Similarly if the complex part of the input is 0 then the complex part of the weight is assigned as -1 or if the complex part of the input is 1 then the weight is also assigned as 1.

7. The weight from the bias neuron to a hidden neuron is assigned as $r - s + 1$, where $r$ is the radius of generalization. The value of $s$ is assigned as sum of the number of ones, $i$'s, and twice the number of *(1+i)*s in the training vector corresponding to the hidden neuron.

8. If the desired output is 0 the output layer weight is set as an inhibitory -1. If the output is 1, then the weight is set as 1.

9. The altered combination procedure of the inputs and the weights causes the hidden neuron inputs to be entirely real. Thus the activation function required at the hidden layer is simply the binary step activation function. The output layer also uses a binary step activation function.

When an input vector is presented to the network, the real and imaginary parts of each input element of the vector are multiplied to the corresponding interconnection weight's real and imaginary parts respectively. The two products are then added to obtain the individual contribution by an input element in the vector. This is done for each element and their individual contributions are then added together to obtain the total contribution of the entire vector. Using this combination procedure, each hidden neuron always receives only real inputs. Thus only a binary step activation function is required at the hidden layer. As an example consider the vector (1 1+i i). The corresponding weight vector is (1-i 1+i -1+i). The input vector now combines with this weight vector to yield a total contribution of 4. This contribution is computed as follows:

$$(Re (1)*Re (1-i) + Im (1)*Im (1-i))$$
$$+ (Re (1+i)*Re (1+i) + Im (1+i)*Im (1+i))$$
$$+ (Re (i)*Re (-1+i) + Im (i)*Im (-1+i)) = 4$$

The 3C algorithm can be expressed as a set of simple **if-then** rules. The formal algorithm is as follows: -

```
for each training vector xm [n] do
    sm = no of 1 s + no of i's + 2*(no of (1+i) s) in xm[1:n-1];
    for index = 1 to n-1 do              // wm [ ]: input weights
        if Re(xm [index]) = 0 then
            Re(wm [index]) = -1;
        end if
        if Re(xm [index]) = 1 then
            Re(wm [index]) = 1;
        end if
        if Im(xm [index]) = 0 then
            Im(wm [index]) = -1;
        end if
        if Im(xm [index]) = 1 then
            Im(wm [index]) = 1;
        end if
    end for
    wm [n] = r - sm + 1;
    for index1 = 1 to k do               // k = no of outputs y
        if ym [index1] = 0 then
            owm [index1] = -1;           // owm [ ]: output wts
        else
            owm [index1] = 1;
        end if
    end for
end for
```

Let $r = 0$, now when an input vector is presented to the network each input neuron receives each element in the vector as input. These inputs combine with their respective weights and all input neurons together, except the bias neuron, provide the hidden neuron corresponding to the input vector with a contribution equal to the *s* value of the vector. And since *r* is set as zero, the contribution from the bias neuron is equal to -*s* + 1. Thus the total input to the hidden neuron is 1. All other hidden neurons receive zero or negative input. This ensures that only one hidden neuron fires for each input.

The following examples illustrate the working of the algorithm. The first example is similar to the XOR function but 0 is replaced by *i* and 1 is replaced by *1+i*. The next example shows how the algorithm works with two output neurons.

**Example 1**

The inputs and outputs are shown below in Table 2. The 3C algorithm can be used to train a network to map these inputs to the outputs. The network architecture is shown in Figure 1 and the various network parameters are tabulated in Table 3.

Table 2: Inputs and outputs for Example 1

| Inputs | | Output |
|---|---|---|
| $X_1$ | $X_2$ | Y |
| i | i | 0 |
| i | 1+i | 1 |
| 1+i | i | 1 |
| 1+i | 1+i | 0 |

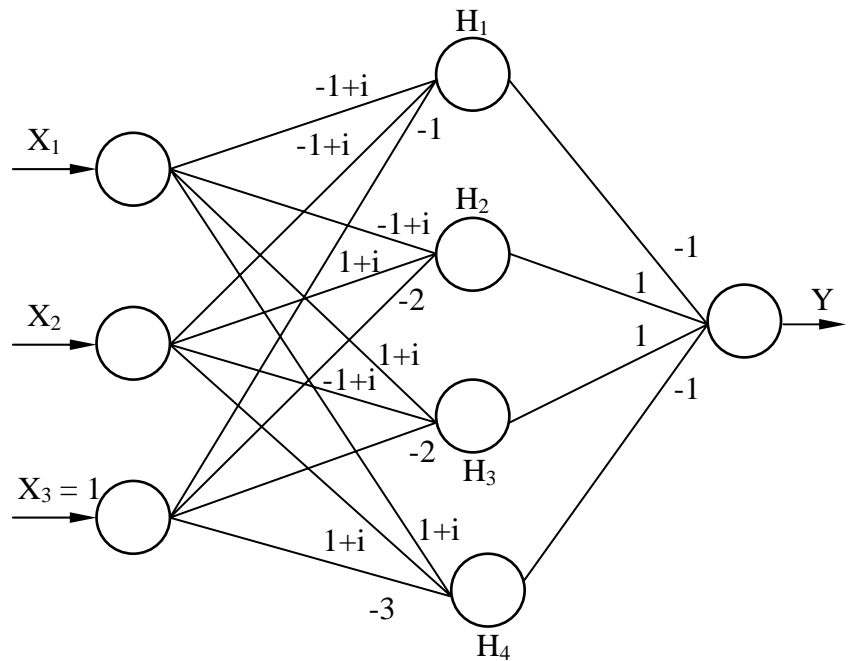

Figure 1: The Network Architecture for Example 1

Table 3: Network Parameters in the input/output mapping of Example 1

| Inputs | | s | Weights | | | Input to H$_1$ | H$_2$ | H$_3$ | H$_4$ | Output of H$_1$ | H$_2$ | H$_3$ | H$_4$ | Output of y |
|---|---|---|---|---|---|---|---|---|---|---|---|---|---|---|
| i | i | 1 | 2 | -1+i | -1+i | -1 | 1 | 0 | 0 | -1 | 1 | 0 | 0 | 0 | 0 |
| i | 1+i | 1 | 3 | -1+i | 1+i | -2 | 0 | 1 | -1 | 0 | 0 | 1 | 0 | 0 | 1 |
| 1+i | i | 1 | 3 | 1+i | -1+i | -2 | 0 | -1 | 1 | 0 | 0 | 0 | 1 | 0 | 1 |
| 1+i | 1+i | 1 | 4 | 1+i | 1+i | -3 | -1 | 0 | 0 | 1 | 0 | 0 | 0 | 1 | 0 |

Each input vector has two elements and so three input neurons are required including the one for the bias neuron. All four samples are required for the training. Thus four hidden neurons are used. The inputs need to be mapped to just one output in each case and so only one output neuron is used here. Also since no generalization is required we have $r = 0$. The weights are assigned according to the algorithm and the network is then tested with all inputs. It is seen that all inputs have been successfully mapped to their outputs.

**Example 2: Network with two output neurons**

The 3C algorithm can also be used to map inputs to a network with more than one output neuron. The inputs and outputs are shown in Table 4. The input vectors have five elements and the corresponding output vectors have two elements.

Table 4: Inputs and outputs for Example 2

| Inputs | | | | | Outputs | |
|---|---|---|---|---|---|---|
| X$_1$ | X$_2$ | X$_3$ | X$_4$ | X$_5$ | Y$_1$ | Y$_2$ |
| 0 | 1+i | 1+i | 0 | i | 1 | 1 |
| 1+i | 0 | 1 | 1+i | 1 | 0 | 1 |
| 1 | 1 | i | 0 | 1 | 1 | 0 |

A total of six input neurons are required including the bias neuron. All three samples need to be used for training and hence three hidden neurons are required. The output layer consists of two neurons. The input and output weights are assigned according to the algorithm as each training sample is presented. No generalization is required so $r = 0$. After the training, the network is tested for all inputs and outputs. Again it can be seen that the mapping is accomplished successfully. The network architecture is shown in Figure 2 and the various network parameters obtained during the training are tabulated in Table 5.

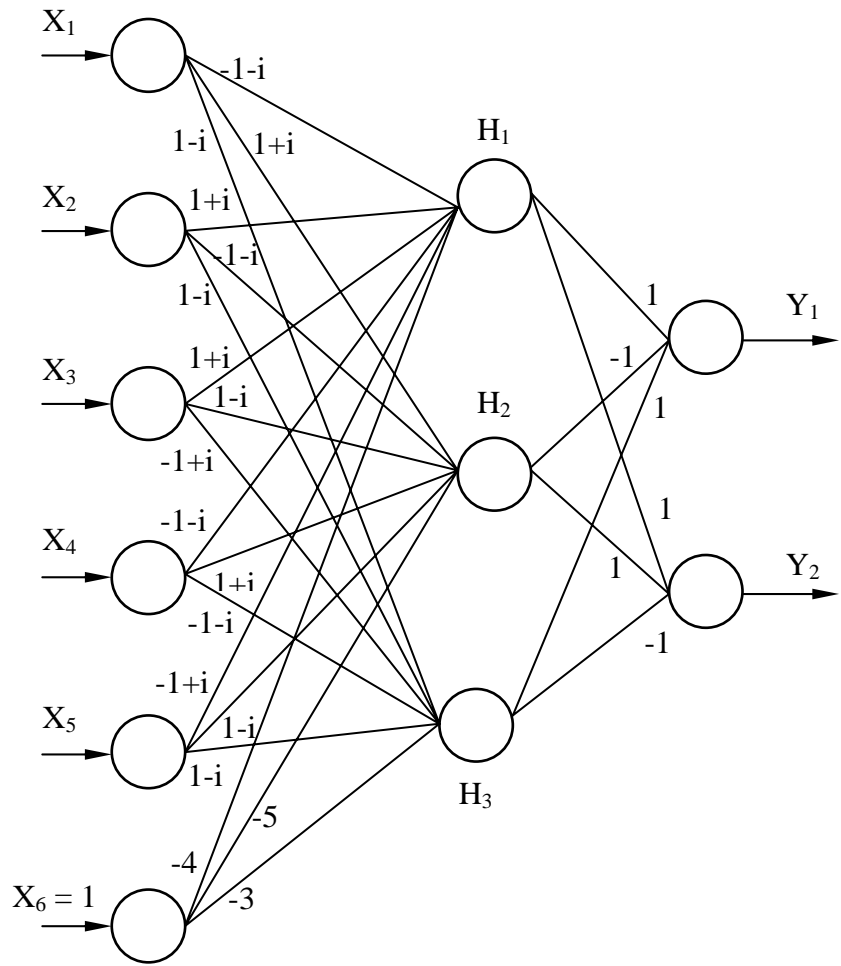

Figure 2: The Network Architecture for Example 2

Table 5: Network Parameters in the input/output mapping of Example 2

| Inputs | | | | | | s | Weights | | | | | |
|---|---|---|---|---|---|---|---|---|---|---|---|---|
| 0 | 1+i | 1+i | 0 | i | 1 | 5 | -1-i | 1+i | 1+i | -1-i | -1+i | -4 |
| 1+i | 0 | 1 | 1+i | 1 | 1 | 6 | 1+i | -1-i | 1-i | 1+i | 1-i | -5 |
| 1 | 1 | i | 0 | 1 | 1 | 4 | 1-i | 1-i | -1+i | -1-i | 1-i | -3 |

| Input to | | | Output of | | | Output | |
|---|---|---|---|---|---|---|---|
| $H_1$ | $H_2$ | $H_3$ | $H_1$ | $H_2$ | $H_3$ | y1 | y2 |
| 1 | -8 | -4 | 1 | 0 | 0 | 1 | 1 |
| -8 | 1 | -5 | 0 | 1 | 0 | 0 | 1 |
| -4 | -5 | 1 | 0 | 0 | 1 | 1 | 0 |

These above examples show how well the network can be trained to store vectors and then associate the vectors with their appropriate outputs when the vectors are presented to the network again. However the generalization property cannot be observed since in both examples $r$ is set to 0. This property of the 3C algorithm can be analyzed by a pattern classification experiment. The algorithm is used to train a network to separate two regions of a spiral pattern. The original pattern is shown in Figure 3 (a). The 16 by 16 area is divided into a black spiral shaped region and another white region. A point in the black spiral region is represented as a binary "1" and a point in the white region is represented by a binary "0". Any point in the region is represented by row and column coordinates. These coordinates, simply row and column numbers, are encoded using 5-character quaternary encoding. These two codes are concatenated and then a bit is added for the bias. This 11-character vector is fed as input to the network. The corresponding outputs are 1 or 0, to denote the region that the point belongs to.

The training samples are randomly selected points from the two regions of the pattern. The samples used here are shown in Figure 3 (b). The points marked "#" are the points from the black region and the points marked "o" are points from the white region. A total of 75 points are used for training. Thus the network used for this pattern classification experiment has 11 neurons in the input layer and 75 neurons in the hidden layer. The output layer requires only one neuron to display a binary "0" or "1".

After the training is done the network is tested for all 256 points in the 16 by 16 area of the pattern. The experiment is repeated then by changing the value of $r$ from 1 to 4. The results for the different levels of generalization achieved are presented in Figure 3 (c), (d), (e) and (f). It can be seen that as the value of $r$ is increased the network tends to generalize more points as belonging to the black region. This over generalization is because during training, the density of the samples presented from the black region was greater than the density of samples from the white region. A summary of the experiment is presented in Table 6. This table contains the number of points classified and misclassified during the testing.

Table 6: No. of points classified/misclassified in the spiral pattern

|   |   | No. of points | | | |
|---|---|---|---|---|---|
|   |   | $r=1$ | $r=2$ | $r=3$ | $r=4$ |
| Spiral Pattern | Classified | 230 | 232 | 233 | 220 |
|   | Misclassified | 26 | 24 | 23 | 36 |

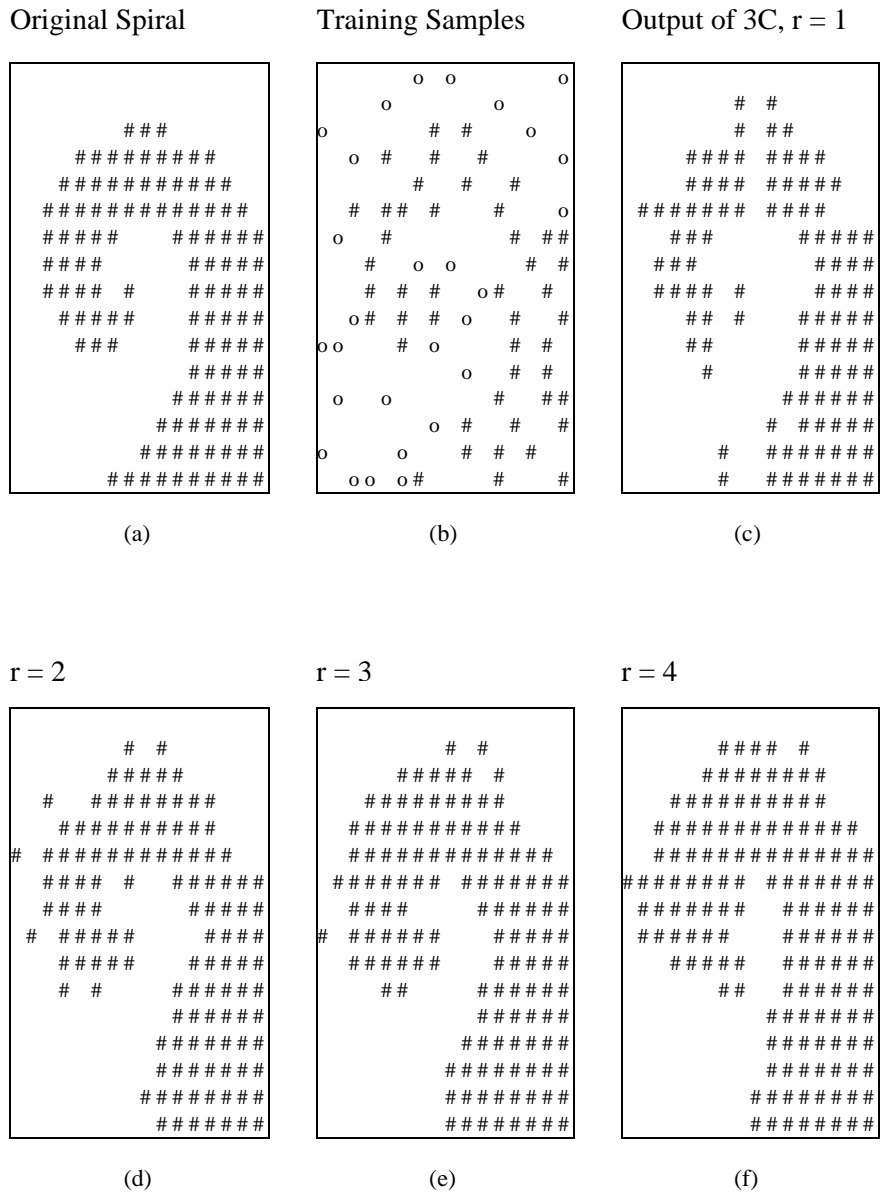

Figure 3: Results of spiral pattern classification

## 4  Time Series Prediction

The Mackey-Glass time series is commonly used to test the performance of neural networks. The series is a chaotic time series making it an ideal representation of the nonlinear oscillations of many physiological processes. The discrete time representation of the series was used earlier to test the performance of the CC4 algorithm. The same will be used here to test the performances of the 3C algorithm.

The discrete time representation of the Mackey-Glass equation is given below:

$$x(k+1) - x(k) = \alpha x(k-\tau) / \{1 + x^\gamma(k-\tau)\} - \beta x(k)$$

The values of the different parameters in the equation are assigned as follows:

$\alpha = 3, \beta = 1.0005, \gamma = 6, \tau = 3$

Since $\tau = 3$, four samples are required to obtain a new point. Thus the series is started with four arbitrary samples: -

$x(1) = 1.5, \quad x(2) = 0.65, \quad x(3) = -0.5, \quad x(4) = -0.7$

Using these samples a series of 200 points is generated and it oscillates within the range -2 to +2. Of these 200 points about nine tenths are fed to the network designed by the 3C algorithm for training. Then the network is tested using the remaining points. In the training and the testing four consecutive points in the series are given as input and the next point is used as the output. Thus a sliding window of size four is used at each and every step. So if nine tenths of the points are to be used for training the total number of sliding windows available is 175, where the first window consists of points 1 to 4 with the 5$^{th}$ point as the output, and the last window consists of points 175 to 178 with the 179$^{th}$ point as output.

The range of the series is divided into 16 equal regions and a point in each region can be represented by the index of the region. These indices ranging from 1 to 16 can be represented using the quaternary encoding scheme. Since four points are required in each training or testing, the 5 character codewords for each of the four inputs are concatenated together. Thus each input vector has 21 elements, where the last element in the vector represents the bias. Unlike the inputs, output points are binary encoded using four bits. This is done to avoid the possibility of generating invalid output vectors that would not belong to the class of expected vectors of the quaternary encoding scheme. Hence 21 neurons are required in the input layer, 175 in the hidden layer (one for each sliding window), and 4 in the output layer.

After the training, the network is tested using the same 175 windows to check its learning ability. Then the rest of the windows are presented to predict future values. The inputs are always points from the original series calculated by the Mackey-Glass equation to avoid an error buildup. The outputs of the network are compared against the expected values in the series. The performance of the 3C algorithm for different values of *r* is presented in the Figures 5, 6, 7 and 8. The values of *r* here are 4, 5, 6 and 7 respectively.

In each of the figures only points 160 to 200 are shown for readability. The solid line represents the original series and the lighter line represents the outputs of the network designed by the 3C algorithm. The lighter line from point 160 to 179 shows how well the network has learnt the samples for different values of *r*. The points predicted by the network are represented by a "×" on the lighter line. The actual points generated by the Mackey-Glass equation are represented by a "o" on the solid line. The first point that is predicted is the point number 180 using the original series points 176, 177, 178 and 179. The next point that is predicted is 181 using the points 177, 178, 179 and 180. The point number 180, which is used as input here, is the original point in the series generated by the Mackey-Glass equation and not the point predicted by the network. Similarly the last point to be predicted is the point number 200 using the actual points 196 to 199 from the series. The network always predicts one point ahead of time and most of the points from 180 to 200 are predicted with very high accuracy. Also the network is able to predict the turning points in the series efficiently. Thus the network is capable of learning the quasi-periodic property of the series. This ability is of great importance in financial applications where predicting the turning point of the price movement is more important than predicting the day to day values.

Stability of networks is another important feature in deciding the network performance and is governed by the consistency of the outputs when network parameters are changed.

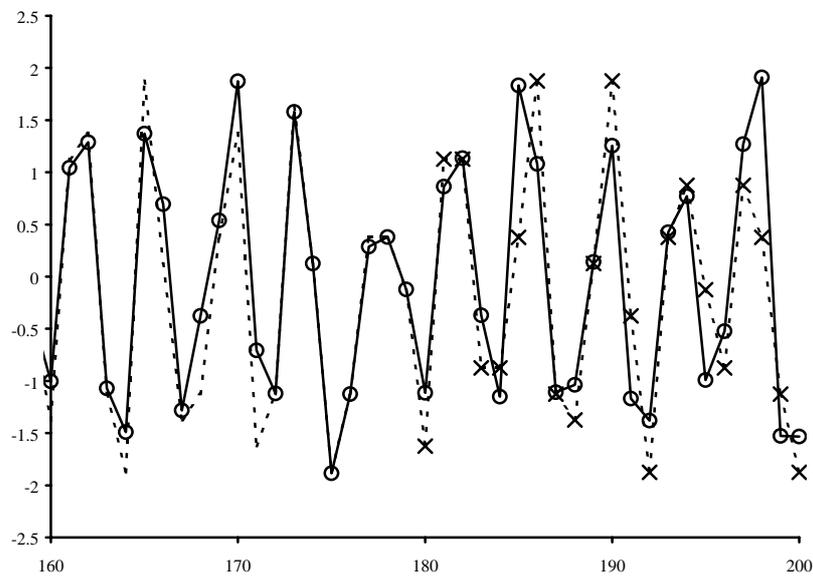

Figure 5: Mackey-Glass time series prediction using 3C, $r = 4$
Dotted line till point 180 – training samples
"o" – Actual data, "×" – predicted data

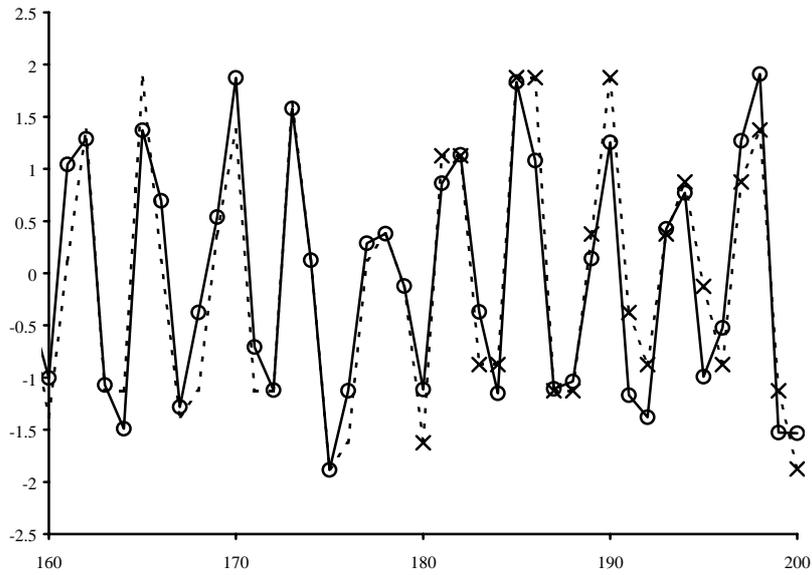

Figure 6: Mackey-Glass time series prediction using 3C, $r = 5$
Dotted line till point 180 – training samples
"o" – Actual data, "×" – predicted data

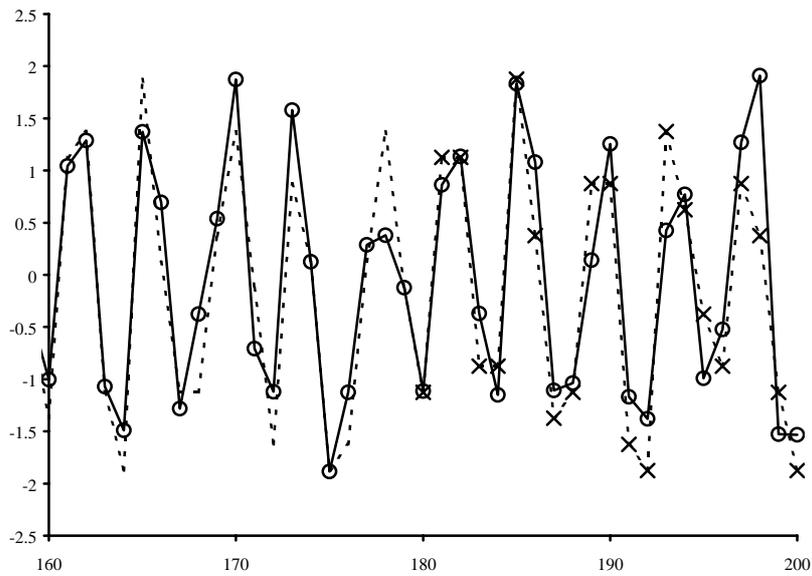

Figure 7: Mackey-Glass time series prediction using 3C, $r = 6$
Dotted line till point 180 – training samples
"o" – Actual data, "×" – predicted data

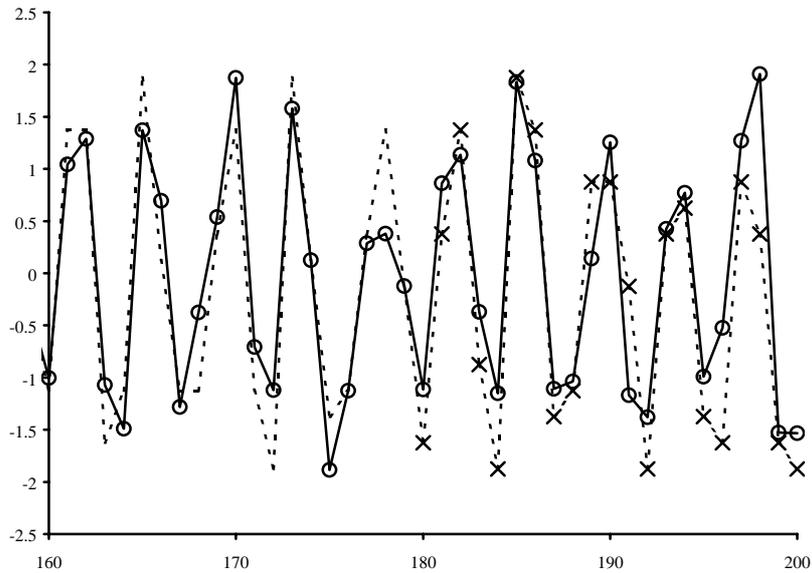

Figure 8: Mackey-Glass time series prediction using 3C, $r = 7$
Dotted line till point 180 – training samples
"o" – Actual data, "×" – predicted data

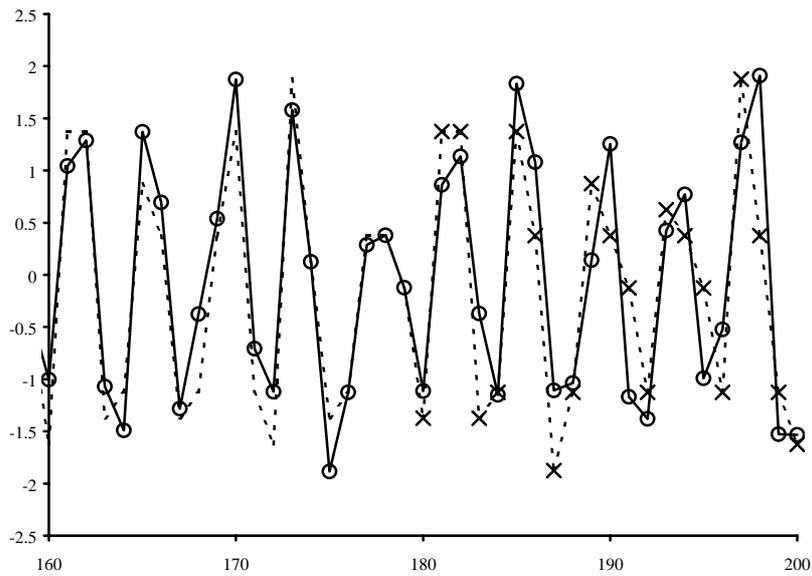

Figure 9: Mackey-Glass time series prediction using 3C, $r = 10$
Dotted line till point 180 – training samples
"o" – Actual data, "×" – predicted data

The use of different values of *r* shows its robustness with regard to generalizability. The normalized mean square error of the points predicted for each value of *r* is shown in Table 7.

Table 7: Normalized mean square error

| Normalized Mean Square Error of predicted points for varying *r* | | | | |
|---|---|---|---|---|
| *r* = 4 | *r* = 5 | *r* = 6 | *r* = 7 | *r* = 10 |
| 0.0238 | 0.0112 | 0.0193 | 0.0221 | 0.0275 |

# 5    Conclusions

Previously, training of complex input neural networks was done using techniques like the backpropagation and perceptron learning rules. These techniques require considerable time and resources to complete the training. The 3C algorithm, which is a generalization of the CC4 algorithm, accomplishes the training instantaneously and requires hardly any resources. Its performance was tested using the pattern classification and time series experiments and its generalization capability was found to be satisfactory.

The quaternary encoding technique makes some modifications to unary encoding so as to accommodate all four characters of the input alphabet.

Like the CC4, the 3C algorithm has its limitations. First of all it can handle only four input values. Also, as with the CC4, a network of the size required by the 3C algorithm poses a problem with respect to hardware implementation. However its suitability for software implementation due to low requirement of computational resources and its instantaneous training make up for the limitations.

In the future the 3C algorithm should be modified and adapted to handle non-binary complex inputs. This would remove the need for encoding the inputs in many cases and greatly increase the number of areas of application. The 3C algorithm can be applied to applications such as financial analysis and communications and signal processing. In most financial applications it is not enough to predict the future price of an equity or commodity; it is more useful to predict when the directionality of price values will change. One could define two modes of behavior, namely the *up* and *down* trends and represent them by the imaginary 0 and 1. Within a trend, the peak and trough could be represented by the real 0 and 1. This gives us four states namely 0, 1, i and 1+i, which can be presented as inputs to the 3C algorithm. In communications and signal processing, the complex inputs of many passband modulation schemes could be directly applied to our feedforward network.

One may also consider even more generalized networks with quaternion inputs. Such networks are likely to provide greater flexibility in the modeling of certain engineering and financial situations. Beyond this remains the task of fitting the family of these networks in the hierarchy of different modes of learning that characterize biological systems.

**References**

1. A. Hirose (ed.) (2003), *Complex-valued Neural Networks: Theories and Applications*, World Scientific Publishing, Singapore.

2. Z. Jihan, P. Sutton (2003), "An FPGA implementation of Kak's instantaneously-trained, fast-classification neural networks" Proceedings of the 2003 IEEE International Conference on Field-Programmable Technology (FPT).

3. S. Kak (1993), "On training feedforward neural networks", *Pramana J. Physics*, vol. 40, pp. 35-42.

4. S. Kak (1994), "New algorithms for training feedforward neural networks", *Pattern Recognition Letters*, vol. 15, pp. 295-298.

5. S. Kak (1998), "On generalization by neural networks", *Information Sciences*, vol. 111, pp. 293-302.

6. S. Kak (2002), "A class of instantaneously trained neural networks", *Information Sciences*, vol. 148, pp. 97-102.

7. S. Kak (2005), "Artificial and biological intelligence", *ACM Ubiquity,* vol. 6, No. 42, pp. 1-20. Also arXiv: cs.AI/0601052

8. A. Ponnath (2006), "Instantaneously trained neural networks", arXiv: cs/0601129

9. P. Raina (1981), "Comparison of learning and generalization capabilities of the Kak and the backpropagation algorithms." vol. 81, pp. 261-274.

10. P. Rajagopal and S. Kak (2003), "Complex valued instantaneously trained neural networks." In *Complex-valued Neural Networks: Theories and Applications*, edited by Akira Hirose, World Scientific Publishing, Singapore.

11. J. Shortt , J. G. Keating, L. Moulinier, C. N. Pannell , "Optical implementation of the Kak neural network" Information Sciences 171, 2005, p.273-287.

12. K.W. Tang and S. Kak (2002), "Fast Classification Networks for Signal Processing", *Circuits Systems Signal Processing*, vol. 21, pp. 207-224.

13. J. Zhu and G. Milne, "Implementing Kak neural networks on a reconfigurable computing platform," In FPL 2000, LNCS 1896, R.W. Hartenstein and H. Gruenbacher (eds.), Springer-Verlag, 2000, p. 260-269.

14. S. Kak, "Three languages of the brain: Quantum, reorganizational, and associative, " In *Learning as Self-Organization,* K. Pribram and J. King, eds., Lawrence Erlbaum, Mahwah, N.J., 1996, pp. 185--219.

15. S. Kak, "Quantum information in a distributed apparatus." Found. Phys. 28, 1998, pp. 1005-1012; arXiv: quant-ph/9804047.

16. S. Kak, "The initialization problem in quantum computing." Found. Phys. 29, pp. 267-279, 1999; arXiv: quant-ph/9805002.

17. S. Kak, "Statistical constraints on state preparation for a quantum computer." Pramana, 57 (2001) 683-688; arXiv: quant-ph/0010109.

18. S. Kak, "General qubit errors cannot be corrected." Inform. Sc., 152, 195-202 (2003); arXiv: quant-ph/0206144.

19. S. Kak, "The information complexity of quantum gates." Int. J. of Theo. Physics, 45, 2006; arXiv: quant-ph/0506013

20. A. Ponnath, "Difficulties in the implementation of quantum computers." arXiv: cs.AR/0602096